\PassOptionsToPackage{unicode}{hyperref}
\PassOptionsToPackage{hyphens}{url}
\documentclass[
]{article}
\usepackage{amsmath,amssymb}
\usepackage{lmodern}
\usepackage[utf8]{inputenc}
\IfFileExists{upquote.sty}{\usepackage{upquote}}{}
\IfFileExists{microtype.sty}{
  \usepackage[]{microtype}
  \UseMicrotypeSet[protrusion]{basicmath} 
}{}
\makeatletter
\@ifundefined{KOMAClassName}{
  \IfFileExists{parskip.sty}{%
    \usepackage{parskip}
  }{
    \setlength{\parindent}{0pt}
    \setlength{\parskip}{6pt plus 2pt minus 1pt}}
}{
  \KOMAoptions{parskip=half}}
\makeatother
\usepackage{xcolor}
\IfFileExists{xurl.sty}{\usepackage{xurl}}{} 
\IfFileExists{bookmark.sty}{\usepackage{bookmark}}{\usepackage{hyperref}}
\hypersetup{
  pdftitle={A Multi-lingual Dataset of Classified Paragraphs from Open Access Scientific Publications},
  pdfkeywords={open science, research data, research software, software
mentions, data mentions, grobid, acknowledgments analysis, clinical
trials mentions, scientific text mining},
  hidelinks,
  pdfcreator={LaTeX via pandoc}}
\usepackage[left=3cm, right=3cm, top=3cm, bottom=3cm]{geometry}
\providecommand{\tightlist}{%
  \setlength{\itemsep}{0pt}\setlength{\parskip}{0pt}}
\ifluatex
  \usepackage{selnolig}  
\fi
\newlength{\cslhangindent}
\newlength{\csllabelwidth}
 {
  \setlength{\parindent}{0pt}
  \ifodd #1 \everypar{\setlength{\hangindent}{\cslhangindent}}\ignorespaces\fi
  \ifnum #2 > 0
  \setlength{\parskip}{#2\baselineskip}
  \fi
 }%
 {}
\usepackage{calc}

\newenvironment{cslreferences}%
  {\setlength{\parindent}{0pt}%
  \everypar{\setlength{\hangindent}{\cslhangindent}}\ignorespaces}%
  {\par}

\title{A Multi-lingual Dataset of Classified Paragraphs from Open Access
Scientific Publications}
\usepackage{authblk}
\author[%
  1%
  ]{%
  Eric Jeangirard%
}
\affil[1]{French Ministry of Higher Education and Research, Paris,
France}
\date{October 2025}

\makeatletter
\def\@maketitle{%
  \newpage \null \vskip 2em
  \begin {center}%
    \let \footnote \thanks
         {\LARGE \@title \par}%
         \vskip 1.5em%
                {\large \lineskip .5em%
                  \begin {tabular}[t]{c}%
                    \@author
                  \end {tabular}\par}%
                                                \vskip 1em{\large \@date}%
  \end {center}%
  \par
  \vskip 1.5em}
\makeatother

\begin{document}
\maketitle

\textbf{Keywords}: open science, research data, research software,
software mentions, data mentions, grobid, acknowledgments analysis,
clinical trials mentions, scientific text mining

\hypertarget{abstract}{%
\subsection{Abstract}\label{abstract}}

We present a dataset of 833k paragraphs extracted from CC-BY licensed
scientific publications, classified into four categories:
acknowledgments, data mentions, software/code mentions, and clinical
trial mentions. The paragraphs are primarily in English and French, with
additional European languages represented. Each paragraph is annotated
with language identification (using fastText) and scientific domain
(from OpenAlex). This dataset, derived from the French Open Science
Monitor corpus and processed using GROBID, enables training of text
classification models and development of named entity recognition
systems for scientific literature mining. The dataset is publicly
available on HuggingFace https://doi.org/10.57967/hf/6679 under a CC-BY
license.

\hypertarget{background-summary}{%
\subsection{Background \& Summary}\label{background-summary}}

Scientific publications contain structured information that extends
beyond the main research findings. Acknowledgments sections reveal
funding sources and research infrastructures; data availability
statements indicate research datasets; software mentions document
computational tools; and clinical trial references provide links to
registered studies. Automatically identifying and extracting these
elements is crucial for research assessment, reproducibility studies,
and understanding the research ecosystem.

While tools like GROBID can identify certain sections (e.g.,
acknowledgments) in English-language publications, coverage for other
languages and paragraph types remains limited. This dataset addresses
this gap by providing labeled paragraphs across multiple categories and
languages, enabling the development and improvement of:

\begin{enumerate}
\def\labelenumi{\arabic{enumi}.}
\tightlist
\item
  \textbf{Text classification models} to identify paragraph types in
  scientific publications, particularly for non-English languages
\item
  \textbf{Named entity recognition (NER) models} to extract:

  \begin{itemize}
  \tightlist
  \item
    Dataset names, DOIs, and accession numbers
  \item
    Software names, repository URLs (GitHub, GitLab), and identifiers
    (SWHID, Wikidata)
  \item
    Clinical trial identifiers (e.g., NCT numbers)
  \item
    Funding agencies, grant IDs, infrastructures, and private entities
    in acknowledgments
  \end{itemize}
\end{enumerate}

The dataset draws from the French Open Science Monitor (FOSM), which
tracks open science practices in France, and leverages existing text
mining tools (GROBID, Softcite, DataStet) for semi-automatic annotation.

\hypertarget{methods}{%
\subsection{Methods}\label{methods}}

\hypertarget{source-corpus}{%
\subsubsection{Source Corpus}\label{source-corpus}}

The source publications were obtained from the French Open Science
Monitor corpus, described in detail by (Bracco et al. 2022). The FOSM
aggregates metadata and full-text content from French-affiliated
scientific publications available under open licenses. Publications are
primarily from the period 2013 onwards, ensuring contemporary research
practices are represented.

\hypertarget{paragraph-extraction}{%
\subsubsection{Paragraph Extraction}\label{paragraph-extraction}}

Paragraphs were extracted from PDF documents using GROBID (GeneRation Of
BIbliographic Data), a machine learning library for extracting and
structuring raw documents into structured XML/TEI formats. GROBID's
document segmentation capabilities identify distinct textual units
within publications, including body paragraphs, acknowledgments, and
supplementary sections.

\hypertarget{classification-and-annotation}{%
\subsubsection{Classification and
Annotation}\label{classification-and-annotation}}

Paragraphs were semi-automatically classified into four categories using
specialized tools:

\begin{enumerate}
\def\labelenumi{\arabic{enumi}.}
\tightlist
\item
  \textbf{Acknowledgments}: Identified using GROBID's native section
  recognition, plus extra heuristics
\item
  \textbf{Data mentions}: Detected using DataStet, a tool for
  identifying dataset mentions in scientific text, plus extra heuristics
\item
  \textbf{Software/Code mentions}: Identified using Softcite, a tool for
  detecting software mentions, plus extra heuristics
\item
  \textbf{Clinical trial mentions}: Detected through pattern matching
  for trial identifiers
\end{enumerate}

Softcite and Datastet are described in (Bassinet et al. 2023), they
machine learning methods for entity recognition in scientific
literature.

\textbf{Note:} No inter-annotator validation was performed. Users should
be aware that classification accuracy depends on the performance of the
underlying tools and may contain errors.

\hypertarget{language-identification}{%
\subsubsection{Language Identification}\label{language-identification}}

Language detection was performed using fastText's language
identification model (lid.176.bin) (Joulin et al. 2016), which supports
176 languages. This pre-trained model provides efficient and accurate
language prediction for short and medium-length texts. This detected
language was kept instead of the language metadata provided by OpenAlex.

\hypertarget{scientific-domain-assignment}{%
\subsubsection{Scientific Domain
Assignment}\label{scientific-domain-assignment}}

Each paragraph was assigned a scientific domain based on the OpenAlex
classification system. Specifically, we used the \texttt{field}
attribute from the \texttt{primary\_topic} of the source publication.
OpenAlex provides a hierarchical classification covering major
scientific disciplines.

\hypertarget{data-structure}{%
\subsubsection{Data Structure}\label{data-structure}}

The dataset is distributed as a CSV file with the following columns:

\begin{itemize}
\tightlist
\item
  \texttt{license}: License, from OpenAlex data
\item
  \texttt{text}: Text content of the paragraph
\item
  \texttt{doi}: DOI of the publication
\item
  \texttt{type}: Publication type, from OpenAlex data
\item
  \texttt{detected\_lang}: ISO 639-1 detected language code, with
  fasttext
\item
  \texttt{publication\_year}: Publication year
\item
  \texttt{is\_dataset}: Boolean, true if the paragraph mentions data
\item
  \texttt{is\_software}: Boolean, true if the paragraph mentions
  software or code
\item
  \texttt{is\_acknowledgement}: Boolean, true if this is an
  acknowledgement paragraph
\item
  \texttt{is\_clinicaltrial}: Boolean, true if the paragraphs mentions a
  clinical trial
\item
  \texttt{field\_name}: Field of the primary topic, from OpenAlex data
\item
  \texttt{field\_id}: OpenAlex id of the field
\end{itemize}

\hypertarget{data-records}{%
\subsection{Data Records}\label{data-records}}

The dataset is publicly available on HuggingFace at:
https://doi.org/10.57967/hf/6679 The dataset contains 833k paragraphs
distributed as follows:

\begin{itemize}
\tightlist
\item
  \textbf{Acknowledgments}: 108k paragraphs
\item
  \textbf{Data mentions}: 570k paragraphs
\item
  \textbf{Software mentions}: 203k paragraphs
\item
  \textbf{Clinical trials}: 8.7k paragraphs
\end{itemize}

\textbf{Language distribution:}

\begin{itemize}
\tightlist
\item
  English: 98.4\%
\item
  French: 1.5\%
\item
  Other European languages: \textasciitilde0.15\%
\end{itemize}

\textbf{License:} CC-BY 4.0

\textbf{Format:} CSV (UTF-8 encoding)

\hypertarget{technical-validation}{%
\subsection{Technical Validation}\label{technical-validation}}

\hypertarget{classification-quality}{%
\subsubsection{Classification Quality}\label{classification-quality}}

As the dataset was produced through semi-automatic annotation without
systematic manual validation, users should consider the following
limitations:

\begin{enumerate}
\def\labelenumi{\arabic{enumi}.}
\tightlist
\item
  \textbf{Tool-dependent accuracy}: Classification quality depends on
  the performance of GROBID, Softcite, and DataStet on the source
  documents
\item
  \textbf{Language bias}: Tools may perform better on English-language
  content than other languages
\item
  \textbf{Domain variability}: Performance may vary across scientific
  fields
\item
  \textbf{Boundary effects}: Paragraph segmentation may occasionally
  split or merge logical units
\end{enumerate}

We recommend users conduct their own validation on a sample of the data
relevant to their specific use case.

\hypertarget{language-identification-quality}{%
\subsubsection{Language Identification
Quality}\label{language-identification-quality}}

FastText's lid.176.bin model has been shown to achieve high accuracy on
well-formed texts. However, scientific texts may contain:

\begin{itemize}
\tightlist
\item
  Mixed-language content (e.g., English terms in French text)
\item
  Technical terminology and symbols
\item
  Short paragraphs with limited context
\end{itemize}

\hypertarget{usage-notes}{%
\subsection{Usage Notes}\label{usage-notes}}

\hypertarget{potential-applications}{%
\subsubsection{Potential Applications}\label{potential-applications}}

\textbf{1. Paragraph Classification Models}

The dataset can be used to train or fine-tune models for identifying
paragraph types in scientific publications. This could be used to extend
and improve upon GROBID's capabilities, particularly for non-English
languages.

\textbf{2. Named Entity Recognition}

Paragraphs could help building training data for NER models targeting:

\begin{itemize}
\tightlist
\item
  \textbf{Data mentions}: Dataset names, DOIs, accession numbers
\item
  \textbf{Software mentions}: Software names, repository URLs (GitHub,
  GitLab), identifiers (SWHID, Wikidata)
\item
  \textbf{Clinical trials}: NCT identifiers and trial registries
\item
  \textbf{Acknowledgments}: Funding agencies, grant IDs, research
  infrastructures, private entities
\end{itemize}

\textbf{3. Research Ecosystem Analysis}

The dataset enables large-scale studies of:

\begin{itemize}
\tightlist
\item
  Data sharing practices across disciplines
\item
  Software usage patterns in research
\item
  Funding acknowledgment practices
\item
  Clinical trial documentation
\end{itemize}

\hypertarget{recommended-practices}{%
\subsubsection{Recommended Practices}\label{recommended-practices}}

\begin{enumerate}
\def\labelenumi{\arabic{enumi}.}
\tightlist
\item
  \textbf{Data splitting}: Ensure publication-level splits (not
  paragraph-level) to avoid data leakage between train/test sets
\item
  \textbf{Multilingual evaluation}: Report performance separately for
  major language groups
\item
  \textbf{Domain-specific testing}: Validate models across different
  scientific fields
\item
  \textbf{Error analysis}: Manual inspection of misclassifications can
  reveal systematic issues
\end{enumerate}

\hypertarget{limitations}{%
\subsubsection{Limitations}\label{limitations}}

\begin{itemize}
\tightlist
\item
  \textbf{No manual validation}: Classification errors from source tools
  are propagated
\item
  \textbf{Imbalanced distribution}: Categories and languages may be
  unevenly represented
\item
  \textbf{Temporal bias}: Predominantly recent publications (2013+) may
  not reflect historical practices
\item
  \textbf{French affiliation bias}: FOSM focuses on French-affiliated
  research, which may not generalize globally
\item
  \textbf{License restriction}: Only CC-BY content included; results may
  not generalize to closed-access literature
\end{itemize}

\hypertarget{references}{%
\section*{References}\label{references}}
\addcontentsline{toc}{section}{References}

\hypertarget{refs}{}
\begin{cslreferences}
\leavevmode\hypertarget{ref-bassinet:hal-04121339}{}%
Bassinet, Aricia, Laetitia Bracco, Anne L'Hôte, Eric Jeangirard, Patrice
Lopez, and Laurent Romary. 2023. ``Large-scale Machine-Learning analysis
of scientific PDF for monitoring the production and the openness of
research data and software in France.''
\url{https://hal.science/hal-04121339}.

\leavevmode\hypertarget{ref-bracco:hal-03651518}{}%
Bracco, Laetitia, Anne L'Hôte, Eric Jeangirard, and Didier Torny. 2022.
``Extending the open monitoring of open science.''
\url{https://hal.science/hal-03651518}.

\leavevmode\hypertarget{ref-joulin_bag_2016}{}%
Joulin, Armand, Edouard Grave, Piotr Bojanowski, and Tomas Mikolov.
2016. ``Bag of Tricks for Efficient Text Classification.''
\emph{arXiv:1607.01759 {[}Cs{]}}, August.
\url{http://arxiv.org/abs/1607.01759}.
\end{cslreferences}

\end{document}